\documentclass[10pt,twocolumn,letterpaper]{article}

\usepackage[pagenumbers]{cvpr} 

\usepackage{comment}
\usepackage{subcaption}
\usepackage{adjustbox}
\usepackage{booktabs}
\usepackage{siunitx}

\usepackage[dvipsnames]{xcolor}

\definecolor{cvprblue}{rgb}{0.21,0.49,0.74}
\usepackage[pagebackref,breaklinks,colorlinks,citecolor=cvprblue]{hyperref}

\def\paperID{65} 
\def\confName{3DV\xspace}
\def\confYear{2026\xspace}

\title{UFV-Splatter: Pose-Free Feed-Forward 3D Gaussian Splatting Adapted to Unfavorable Views}

\author{Yuki Fujimura$^1$ 
\quad 
Takahiro Kushida$^2$ 
\quad
Kazuya Kitano$^1$ 
\quad 
Takuya Funatomi$^{1,3}$
\quad 
Yasuhiro Mukaigawa$^1$ \\
$^1$NAIST, Japan\quad
$^2$Ritsumeikan University, Japan\quad $^3$Kyoto University, Japan\\
{\tt\small \{fujimura.yuki,kitano.kazuya,funatomi,mukaigawa\}@is.naist.jp}
\quad {\tt\small
tkushida@fc.ritsumei.ac.jp}
}

\begin{document}
\maketitle
\begin{abstract}
This paper presents a pose-free, feed-forward 3D Gaussian Splatting (3DGS) framework designed to handle unfavorable input views. A common rendering setup for training feed-forward approaches places a 3D object at the world origin and renders it from cameras pointed toward the origin---i.e., from favorable views, limiting the applicability of these models to real-world scenarios involving varying and unknown camera poses.
To overcome this limitation, we introduce a novel adaptation framework that enables pretrained pose-free feed-forward 3DGS models to handle unfavorable views.
We leverage priors learned from favorable images by feeding recentered images into a pretrained model augmented with low-rank adaptation (LoRA) layers.
We further propose a Gaussian adapter module to enhance the geometric consistency of the Gaussians derived from the recentered inputs, along with a Gaussian alignment method to render accurate target views for training. 
Additionally, we introduce a new training strategy that utilizes an off-the-shelf dataset composed solely of favorable images.
Experimental results on both synthetic images from the Google Scanned Objects dataset and real images from the OmniObject3D dataset validate the effectiveness of our method in handling unfavorable input views. 
Project page: \url{https://yfujimura.github.io/UFV-Splatter_page/}
\end{abstract}
\section{Introduction}\label{sec:introduction}
3D Gaussian splatting (3DGS)~\cite{Kerbl2023} is a powerful technique for representing 3D scenes using volumetric Gaussians. These Gaussians are optimized to reconstruct multi-view input images given known camera poses. Once optimized, they enable photorealistic novel view synthesis via an efficient differentiable rasterizer. However, estimating both accurate 3D Gaussians and camera poses becomes challenging in scenarios with only sparse input views, motivating the need to address pose-free sparse-view scenarios.

One promising approach to handling sparse input views is to leverage learned scene priors~\cite{Jain2021,Roessle2022,Wang2023,Li2024,Yang2024,Kong2025}, or to employ learning-based feed-forward models~\cite{Yu2021,Chen2021,Charatan2024,Chen2024,Chen2024mvsplat360,Kai2024}. Feed-forward 3DGS models take sparse input views and directly output pixel-aligned 3D Gaussians, offering both strong priors and fast inference without the need for iterative optimization. Recently, pose-free feed-forward 3DGS models~\cite{Xu2024,Ye2025,Zhang2025} have been introduced, where a Vision Transformer (ViT)~\cite{Dosovitskiy2021} predicts pixel-aligned Gaussians without relying on extrinsic or even intrinsic camera parameters.

\begin{figure}[t]
  \centering
    \includegraphics[width=1.\linewidth]{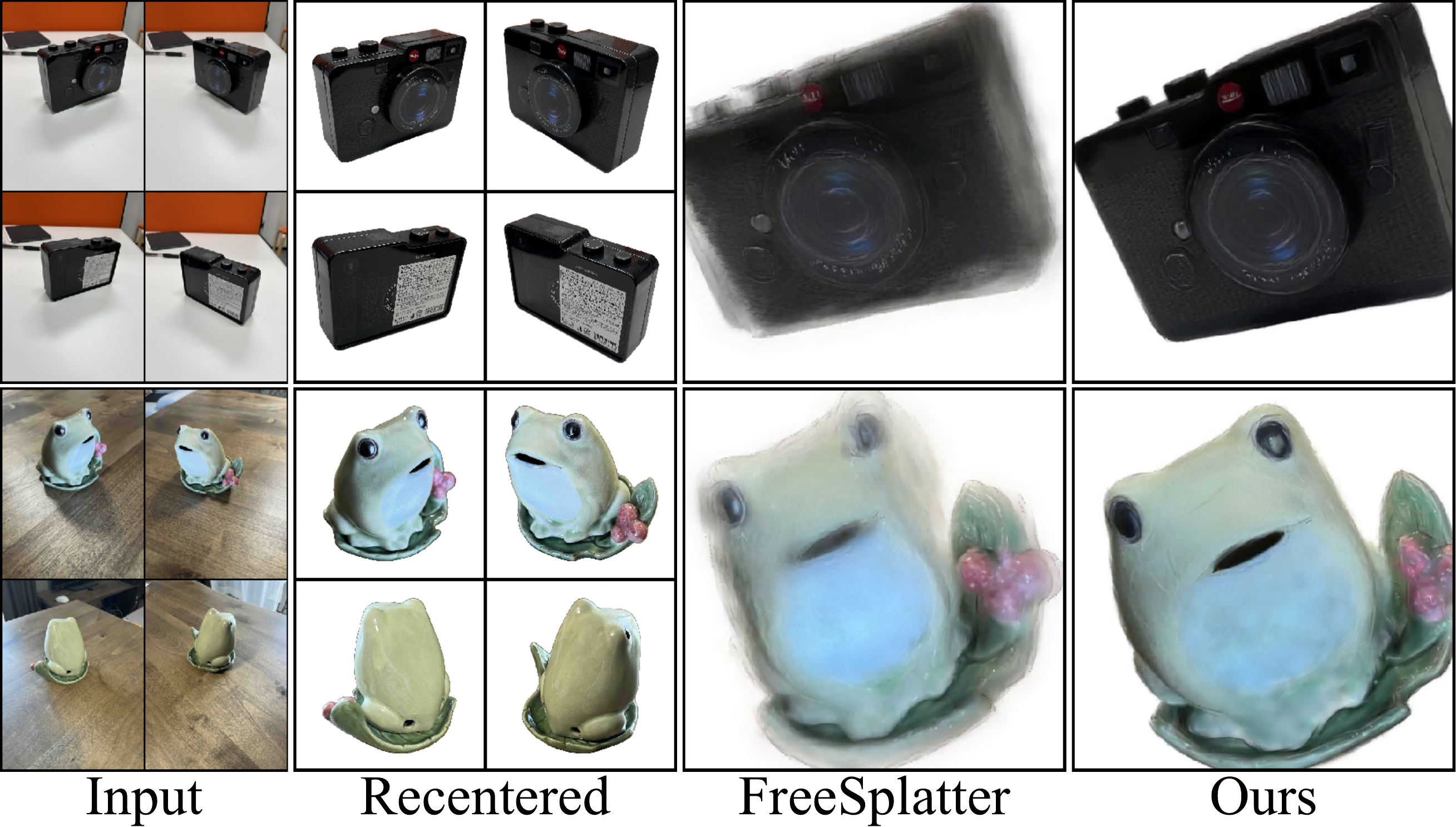}
  \caption{Novel-view synthesis for daily captures with a smartphone. The foreground regions are masked~\cite{Zheng2024} and recentered prior to inference. These results demonstrate the practical applicability of our method in real-world scenarios with unknown and varying camera poses.}
  \label{fig:result_daily}
\end{figure}

\begin{figure*}[t]
  \centering
  \includegraphics[width=1.\linewidth]{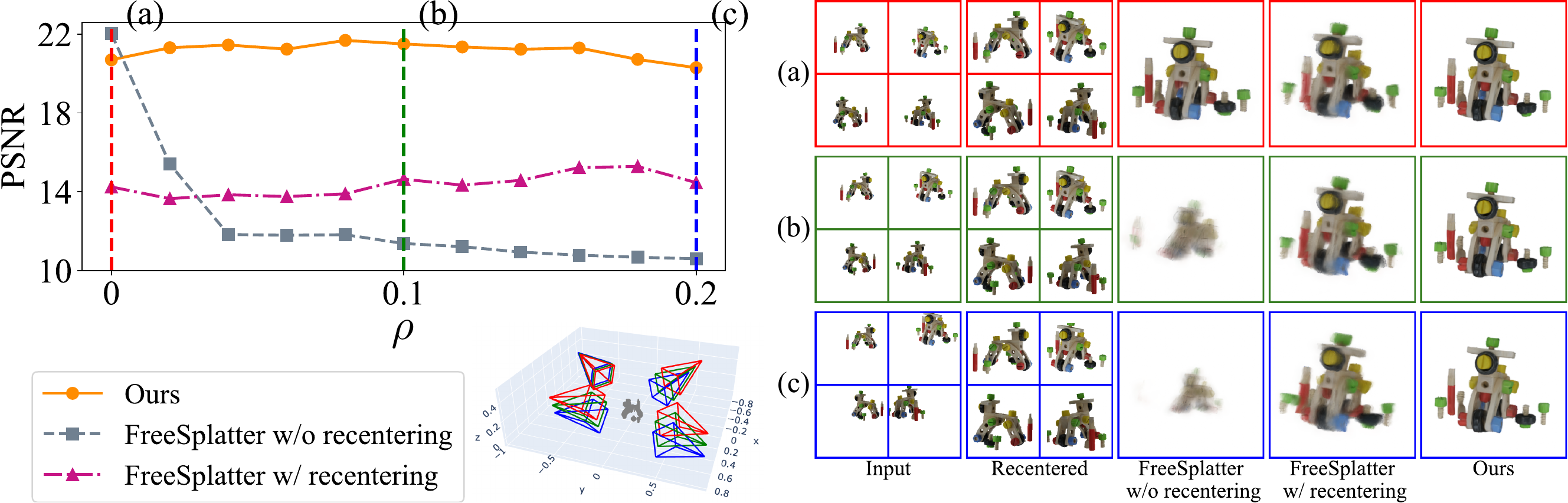}
  \caption{
\textit{Favorable} views (red cameras, $\rho=0$): the object is placed at the world origin, and the cameras are oriented toward the origin. 
\textit{Unfavorable} views (green and blue cameras, $\rho=0.1, 0.2$): generated by adding a translation of random direction and magnitude $\rho r$ to each favorable camera, where $r$ denotes the distance from the object origin to the camera.
FreeSplatter~\cite{Xu2024}, trained only on favorable views, performs well when inputs are also favorable but struggles under unfavorable inputs (FreeSplatter w/o recentering).
A naive approach, such as recentering the foreground to mimic favorable views, is insufficient to resolve these generalization issues (FreeSplatter w/ recentering).
Our proposed method remains robust to unknown and varying camera poses.
}
  \label{fig:challenges}
\end{figure*}

To train such models for object-centric scenes, large-scale synthetic datasets such as Objaverse~\cite{Deitke2023} are commonly used. The multi-view images for training are rendered from millions of 3D assets, typically with the object centered at the world origin and cameras pointing toward the origin---referred to as {\it favorable} views in this study (red cameras in Fig. \ref{fig:challenges}). While this setup facilitates stable training, it limits the generalization ability of pose-free models to real-world scenarios with varying camera poses. Generalizing to such {\it unfavorable} views (green and blue cameras in Fig. \ref{fig:challenges}) remains a significant challenge, often resulting in inconsistent Gaussian predictions across input views.


In this work, we propose a novel framework to adapt a pretrained pose-free feed-forward 3DGS model to handle unfavorable views without requiring new training datasets.
Our method adapts the pretrained model using low-rank adaptation (LoRA)~\cite{Hu2022}.
To leverage the prior learned from favorable views, we recenter the images via simple image shifting and resizing.
We further propose a Gaussian adapter module that improves geometric consistency in the Gaussians derived from the recentered images, along with a Gaussian alignment method that aligns the predictions for accurate target view rendering during training.
Additionally, we introduce a training strategy that eliminates the need for datasets containing unfavorable views. By leveraging the pretrained model to generate 3D Gaussians for each sample in a high-quality off-the-shelf dataset containing only favorable images (e.g., the G-Objaverse dataset~\cite{Zuo2024}), we synthesize training images from unfavorable viewpoints. This enables effective model adaptation without requiring new data creation.
As shown in Fig. \ref{fig:result_daily}, our method enhances the practical applicability of pose-free 3DGS models in real-world scenarios with unknown and varying camera poses.    
\section{Related work}
\paragraph{Novel-view synthesis from sparse views.}
Novel-view synthesis aims to generate images from unseen viewpoints based on observed multi-view images. Neural radiance fields (NeRF)~\cite{Mildenhall2020} and 3D Gaussian splatting (3DGS)~\cite{Kerbl2023} have emerged as powerful techniques in this area. However, these methods typically require dozens to hundreds of input images to produce photorealistic results.

To address the sparse-view scenario, several works incorporate additional cues such as sparse depth from structure-from-motion~\cite{Deng2022}, monocular depth estimation~\cite{Song2023,Wang2023,Uy2023,Chung2024,Li2024,Zhu2024}, view-consistency constraints~\cite{Kwak2023,Seo2023,Lao2023,Truong2023,Mai2024,Zheng2025}, generative priors~\cite{Wynn2023,Yang2024,Zhao2024,Kong2025}, scene-level regularization~\cite{Kim2022,Niemeyer2022}, improved learning strategies~\cite{Yang2023,Zhang2024}, and architectural innovations~\cite{Zhu2024CVPR,Lin2025}. While these approaches enable effective synthesis under sparse views, most assume known or approximately estimated camera poses~\cite{Truong2023,Fan2024,Mai2024}, which are often difficult to obtain, especially when views are limited and camera baselines are large.

\paragraph{Feed-forward models for novel-view synthesis.}
Another promising direction is the use of feed-forward models that directly estimate neural representations~\cite{Chen2021,Yu2021} or 3D Gaussians~\cite{Charatan2024,Chen2024lara,Chen2024,Chen2024mvsplat360,Tang2024,Kai2024} from sparse input views. These models leverage scene priors learned from large-scale datasets to compensate for the lack of dense observations and also enable fast inference without the need for iterative optimization.

Feed-forward 3DGS models commonly aim to estimate pixel-aligned 3D Gaussians. For instance, pixelSplat~\cite{Charatan2024} employs probabilistic Gaussian generation along camera rays to ensure differentiability. MVSplat~\cite{Chen2024} utilizes plane-sweep volumes inspired by multi-view stereo~\cite{Yao2018} to estimate Gaussian centers. GS-LRM~\cite{Kai2024}, built upon the Large Reconstruction Model (LRM)~\cite{Hong2024}, processes tokenized image features via self-attention to capture geometric and photometric consistency.
However, these approaches still rely on known camera poses, which are typically used to constrain Gaussians along camera rays or encoded as model inputs, limiting their applicability to pose-free scenarios.

\paragraph{Pose-free feed-forward models.}
Pose-free feed-forward models aim to estimate neural representations or 3D Gaussians directly from images, without requiring extrinsic or even intrinsic camera parameters. Recent works such as DUSt3R~\cite{Wang2024dust3r} and MASt3R~\cite{Leroy2024} propose foundation models that infer 3D point maps from pose-free image pairs, enabling pose estimation via PnP solvers~\cite{Hartley2003}.
Inspired by these advances, Splatt3R~\cite{Smart2024} and NoPoSplat~\cite{Ye2025} introduce regression heads to estimate Gaussian parameters for novel view synthesis from pose-free image pairs. MV-DUSt3R+~\cite{Tang2025} extends DUSt3R to multi-view settings and incorporates Gaussian parameter estimation for view synthesis.
Another line of research, such as FLARE~\cite{Zhang2025} and PF3splat~\cite{Hong2025}, proposes to use initial poses from a pretrained pose estimator, which are then updated along with Gaussian estimation.

For sparse-view object-centric scenes with large baselines, FORGE~\cite{Jiang2024} sequentially estimates camera poses and neural volumes. LEAP~\cite{Jiang2024LEAP} directly regresses neural representations from pose-free sparse views. PF-LRM~\cite{Wang2024} extends LRM~\cite{Hong2024} to estimate point clouds for PnP-based pose estimation and neural representations from multi-view images. Upfusion~\cite{Nagoor2024} proposes a diffusion model conditioned on input views and target rays to generate novel-view image, although it requires optimization to generate a 3D representation.

SpaRP~\cite{Xu2024_sparp} utilizes Stable Diffusion~\cite{Rombach2022} to produce normalized object coordinate maps from tiled views for pose estimation. It also generates tiled views from uniformly distributed poses, which are then processed by a feed-forward model to predict signed distance fields and colors for novel-view synthesis and 3D reconstruction.
LucidFusion~\cite{He2025} proposes a relative coordinate map, similar to the normalized object coordinate map in SpaRP, to represent scene geometry. The relative coordinate map is then concatenated with Gaussian parameters for novel-view synthesis.

FreeSplatter~\cite{Xu2024} extends GS-LRM~\cite{Kai2024} to estimate pixel-aligned 3D Gaussians from pose-free sparse views. 
Compared to neural-volume-based methods~\cite{Wang2024}, pixel-aligned Gaussians can better reconstruct the fine details of a target object. 
Additionally, FreeSplatter addresses occlusion in pixel-aligned Gaussians by freely moving Gaussians in background regions and compensating for occluded areas.

Our work builds upon these advances by investigating and enhancing the real-world applicability of pose-free feed-forward 3DGS models trained on synthetic data with limited pose variations. Specifically, we focus on adapting such models to operate effectively under unfavorable camera poses, which are common in real-world applications.

\begin{figure*}[t]
\centering
\includegraphics[width=1.0\textwidth]{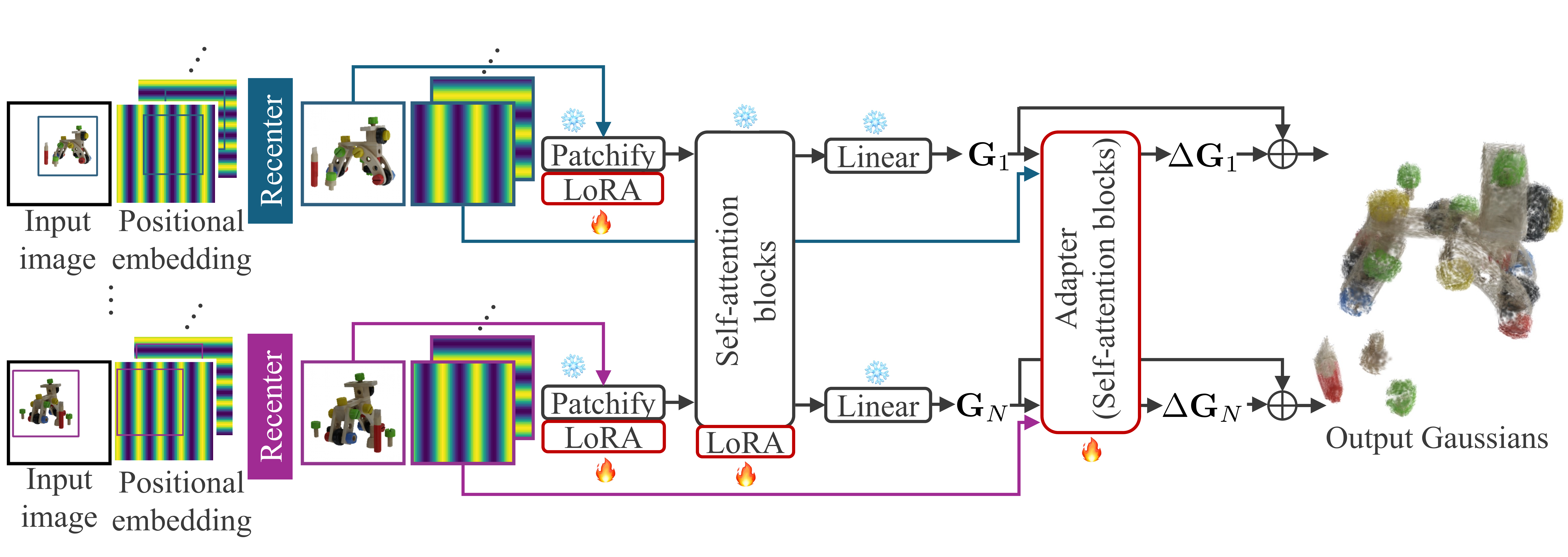}
\caption{Overview of the proposed model. It is built upon a pretrained pose-free feed-forward 3DGS model, augmented with LoRA layers. The model processes recentered input images to leverage the priors learned from favorable views. An additional adapter module corrects geometric errors in the estimated Gaussians caused by the recentering operation, using the recentered 2D positional embeddings.
}
\label{fig:overview}
\end{figure*}

\section{Method}

We aim to adapt a pose-free feed-forward 3DGS model to handle unfavorable input views. We begin by reviewing the formulation of pose-free feed-forward 3DGS (Sec.~\ref{sec:review_pf_model}) and defining unfavorable views (Sec.~\ref{sec:define_ufv}). We then describe our model architecture, which includes an adapter module for geometric refinement (Sec.~\ref{sec:model}). Finally, we present our training strategy, which enables effective learning using only favorable-view images, along with a Gaussian alignment procedure for accurate supervision (Sec.~\ref{sec:training}).

\subsection{Pose-free Feed-forward 3D Gaussian Splatting}\label{sec:review_pf_model}

We first review pose-free feed-forward 3DGS. Given a set of input context images $\{\mathbf{I}_i^c\}_{i=1}^{N},$ where each $\mathbf{I}_i^c \in \mathbb{R}^{H \times W \times 3}$, the task is to generate 3D Gaussians for each image without access to camera poses. Here, $N$ denotes the number of input images, and $H$ and $W$ represent the height and width, respectively. A pose-free feed-forward model $\mathcal{F}$ predicts pixel-aligned 3D Gaussians as follows:

\begin{equation}
\{\mathbf{G}_i\}_{i=1}^{N} = \mathcal{F}(\{\mathbf{I}_i^c\}_{i=1}^{N}), \label{eq:base_eq}
\end{equation}
where each output $\mathbf{G}_i \in \mathbb{R}^{H \times W \times (11+3(d+1)^2)}$ encodes the parameters of the predicted 3D Gaussians, including the Gaussian center $\boldsymbol{\mu}_i \in \mathbb{R}^{H \times W \times 3}$, opacity $\boldsymbol{\alpha}_i \in \mathbb{R}^{H \times W \times 1}$, rotation as quaternions $\mathbf{r}_i \in \mathbb{R}^{H \times W \times 4}$, scale $\mathbf{s}_i \in \mathbb{R}^{H \times W \times 3}$, and view-dependent color $\mathbf{c}_i \in \mathbb{R}^{H \times W \times 3(d+1)^2}$ using spherical harmonics of degree $d$.
During training, the estimated Gaussians are used to render a target image $\mathbf{I}^t \in \mathbb{R}^{H \times W \times 3}$ for computing the training loss.

\subsection{Definition of Unfavorable View}\label{sec:define_ufv}

As illustrated in Fig.~\ref{fig:challenges}, we define \textit{favorable} views as camera poses oriented toward the object center (i.e., the world origin), and \textit{unfavorable} views as those that deviate from this configuration.
For synthetic data, we generate unfavorable views by applying translations to the favorable camera positions. As shown in Fig.~\ref{fig:example_images} in the supplementary material, this results in a wide variety of viewpoints that reflect realistic variations in camera placement.

In Fig.~\ref{fig:challenges}, we construct unfavorable views by adding a translation of random direction and magnitude $\rho r$ to each favorable camera, where $r$ denotes the distance from the object origin to the camera. When $\rho = 0$ (favorable views), the pretrained model $\mathcal{F}$—which is trained solely on such views—can synthesize novel views effectively.
However, when $\rho > 0$ (unfavorable views), $\mathcal{F}$ often produces inconsistent Gaussians across input views, leading to degraded novel-view synthesis quality. Our goal is to adapt $\mathcal{F}$ so that it generates accurate and consistent Gaussians even from unfavorable inputs.
As demonstrated in Fig.~\ref{fig:challenges}, our method is robust to large variations in camera poses.

\subsection{Model Architecture}\label{sec:model}

Given a pose-free feed-forward 3DGS pretrained on favorable views, we aim to adapt it to handle unfavorable views and produce consistent Gaussians across viewpoints. Figure~\ref{fig:overview} provides an overview of the proposed model.

\paragraph{Recentering input images and positional embeddings.}
Directly feeding unfavorable views into the pretrained model leads to significant geometric errors in Gaussian estimation due to domain discrepancy (see ``FreeSplatter w/o recentering'' in Fig. \ref{fig:challenges}).
To leverage the priors learned from favorable views, we recenter each input image by shifting the foreground region to the image center and resizing it (``Recentered'' in Fig. \ref{fig:challenges}). Formally, we define the recentered image as:
\begin{equation}
\overline{\mathbf{I}}_i^c = \mathcal{R}_{\theta_i}(\mathbf{I}_i^c),
\end{equation}
where $\mathcal{R}$ denotes the recentering function, and $\theta_i \in \mathbb{R}^4$ represents the bounding box coordinates—specifically, the pixel positions of the top-left and bottom-right corners. The foreground is resized such that $\max(H_f,W_f) = 0.8 \min(H,W)$, where $H_f$ and $W_f$ are the height and width of the resized foreground bounding box, respectively.
Although this recentering disrupts the geometric correspondence between image pixels and world coordinates—making camera pose recovery via PnP~\cite{Hartley2003} infeasible—it enables the model to exploit the priors from favorable views and facilitates the Gaussian alignment described in Section~\ref{sec:training}.

We also recenter the 2D positional embedding $\mathbf{e}\in \mathbb{R}^{H/p \times W/p \times D}$, used in subsequent modules, as:
\begin{equation}
\overline{\mathbf{e}}_i = \mathcal{R}_{\theta_i/p}(\mathbf{e}),
\end{equation}
where $p$ is a patch size and $D$ is a channel dimension. This recentered embedding reflects the original pixel locations of the foreground region after recentering and is used by the adapter module.

\paragraph{Adapting pose-free feed-forward 3DGS to unfavorable views.}
We adopt FreeSplatter~\cite{Xu2024} as the pose-free feed-forward 3DGS backbone. FreeSplatter encodes patchified input images using a ViT with view-embedding tokens to distinguish the canonical view from the others. These tokens are processed by a linear head to predict pixel-aligned 3D Gaussians, whose coordinates are aligned with the canonical view, following the conventions of~\cite{Leroy2024,Wang2024dust3r}.

To adapt this model to unfavorable views, we insert LoRA layers~\cite{Hu2022} into the patchifying convolution layer and all linear layers within the self-attention blocks (Fig.~\ref{fig:overview}). This allows the model to learn to process recentered images without full model retraining. Please refer to the appendix for comparisons of different LoRA insertion strategies.

\begin{figure*}[t]
\centering
\includegraphics[width=1.0\textwidth]{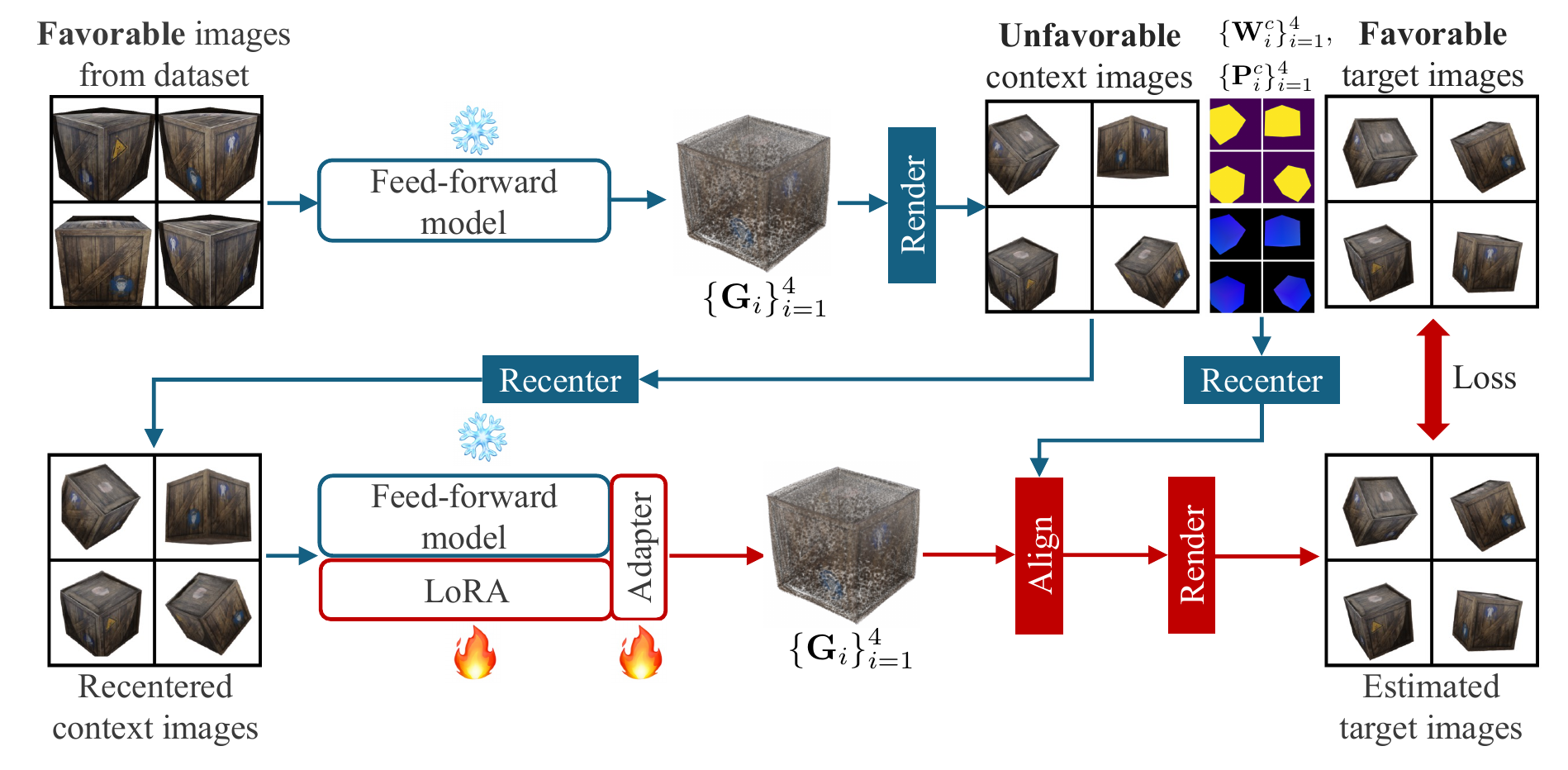}
\caption{Overview of the training framework without requiring an unfavorable-view dataset. Favorable images from an off-the-shelf dataset are fed into a pretrained pose-free model to estimate 3D Gaussians, which are then used to render synthetic unfavorable context views and favorable target views. The recentered context views are processed by the model augmented with LoRA layers and the adapter to predict Gaussians, which are subsequently aligned using recentered opacity and point maps. The aligned Gaussians are finally used to render the target views for computing the training loss.}
\label{fig:training}
\end{figure*}

\paragraph{Adapter for generating geometrically-accurate Gaussians.}
Although the LoRA-adapted model improves cross-view consistency, the geometry of the Gaussians derived from recentered images remains inaccurate. This is because feed-forward models predict pixel-aligned Gaussians, and recentering distorts pixel-to-ray correspondence.

To address this issue, we propose an adapter module that refines the initial Gaussians by predicting residual corrections. Specifically, the Gaussian features $\mathbf{G}_i \in \mathbb{R}^{H \times W \times (11+3(d+1)^2)}$ are first reshaped into patch-sized regions of shape $H/p \times W/p \times p^2 (11+3(d+1)^2)$. These are flattened and projected into tokens $\mathbf{t}_i \in \mathbb{R}^{H/p \times W/p \times D}$ via a linear layer. The tokens are then summed with the recentered positional embedding $\overline{\mathbf{e}}_i$ to encode spatial context.
The resulting tokens from all views are concatenated and processed through multi-head self-attention blocks, enabling the adapter to learn photometric and geometric consistency across views. Finally, a linear head predicts residuals $\Delta \mathbf{G}_i \in \mathbb{R}^{H \times W \times (11+3(d+1)^2)}$, which are added to the original Gaussians to produce the refined output $\widetilde{\mathbf{G}}_i$. Please see the appendix for the details of the architecture and Gaussian update.

\begin{figure}[t]
  \centering
  \subfloat[]{\includegraphics[width=0.23\linewidth]{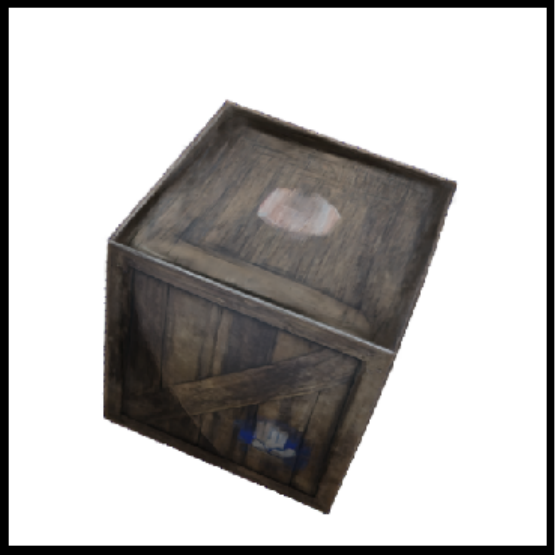}}
  \,
  \subfloat[]
  {\includegraphics[width=0.23\linewidth]{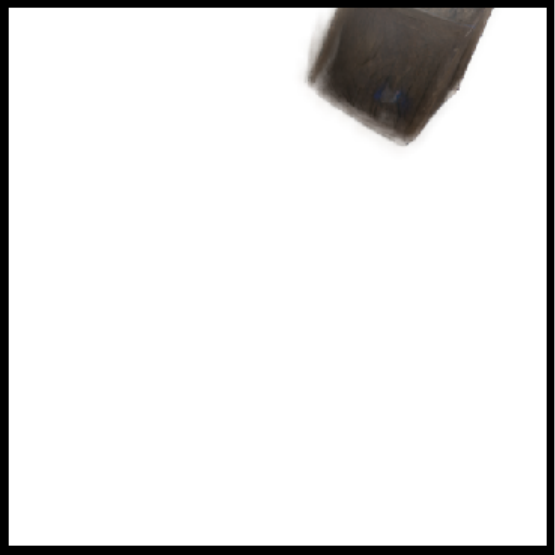}}
  \,
  \subfloat[]{\includegraphics[width=0.23\linewidth]{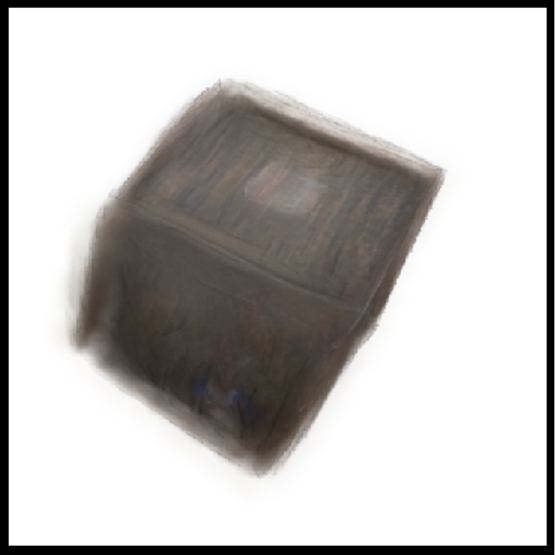}}
  \,
  \subfloat[]{\includegraphics[width=0.23\linewidth]{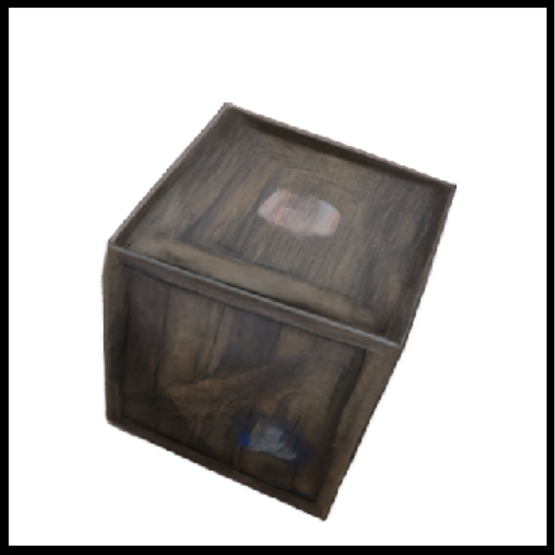}}
  \caption{Effect of Gaussian alignment during training.
(a) Ground-truth target image.
(b) Rendering from unaligned 3D Gaussians at the beginning of training.
(c) Rendering after applying the proposed alignment method.
(d) Rendering from aligned 3D Gaussians after training.
Because of the pose-free input, the estimated Gaussians may vary in scale and FoV across views, resulting in misaligned renderings (b). Our alignment method resolves this inconsistency, enabling accurate supervision through rendered targets.}
  \label{fig:alignement}
\end{figure}

\subsection{Training}\label{sec:training}

We now describe a training framework that does not require a dataset with unfavorable views. Figure~\ref{fig:training} illustrates the overall pipeline.

\paragraph{Gaussian labeling for rendering unfavorable views.}

To avoid the costly process of downloading and filtering large-scale 3D assets to generate unfavorable views, we use an off-the-shelf high-quality 2D dataset such as G-Objaverse~\cite{Zuo2024}, which contains only favorable views. We estimate Gaussians using the pretrained model and synthesize training images from unfavorable viewpoints.

In practice, four favorable views from the dataset are input to the model to estimate the 3D Gaussians. From these, we render $N$ target views $\{\mathbf{I}_i^t\}_{i=1}^N$ and $N$ context views $\{\mathbf{I}_i^c\}_{i=1}^N$.
The target views are rendered by sampling camera positions spherically around the centroid of the estimated Gaussians. For context views, we additionally apply camera translations to simulate unfavorable viewpoints. 
To further encourage robustness, the camera field-of-view (FoV) is also randomly varied.
We also render corresponding opacity maps $\{\mathbf{W}_i^c\}_{i=1}^N$ and depth maps, which are converted into 3D point maps $\{\mathbf{P}_i^c\}_{i=1}^N$ using known intrinsics and extrinsics.
See the appendix for rendering parameters and image samples.

\paragraph{Gaussian alignment for rendering target views.}

To compute loss, we render target images from the estimated Gaussians. However, due to pose-free input and recentering, the predicted Gaussians may differ in scale and FoV across views, and their poses no longer match the camera settings of the target view. As shown in Fig.~\ref{fig:alignement}(b), rendering without correction leads to misaligned results.

To address this, we propose a simple yet effective alignment of the refined 3D Gaussians $\{\widetilde{\mathbf{G}}_i\}_{i=1}^N$.
Specifically, we estimate a scale factor $a \in \mathbb{R}$ and a translation vector $\mathbf{b} \in \mathbb{R}^3$ that minimize the weighted least squares distance between the refined Gaussian centers and the shifted point clouds:
\begin{gather}
\min_{a,\mathbf{b}} \sum_{i=1}^{N} \sum_{q \in \Omega} \overline{\mathbf{W}}_i^c(q)\left\| a
\widetilde{\boldsymbol{\mu}}_i(q)
+ \mathbf{b} - \overline{\mathbf{P}}_i^c(q)
\right\|^2,\\
\Omega = [1,W] \times [1,H],
\end{gather}
where $\widetilde{\boldsymbol{\mu}}_i(q) \in \mathbb{R}^3$ is the center of the refined Gaussian at pixel $q$, and $\overline{\mathbf{P}}_i^c(q)$ is the corresponding 3D point. Here, $\overline{\mathbf{P}}_i^c(q)$ is obtained by applying the recentering operations to the original point map $\mathbf{P}_i^c$ similar to the input image $\overline{\mathbf{I}}_i^c$ as
$\overline{\mathbf{P}}_i^c = \mathcal{R}_{\theta_i}(\mathbf{P}_i^c)$,
thereby ensuring pixel-level correspondence between $\widetilde{\boldsymbol{\mu}}_i(q)$ and $\overline{\mathbf{P}}_i^c(q)$.
We use a similarly recentered opacity map $\overline{\mathbf{W}}_i^c = \mathcal{R}_{\theta_i}(\mathbf{W}_i^c)$ as a weight to focus the alignment on foreground pixels.
This weighted least squares problem can be efficiently solved in closed form during training (please see the appendix for details).
The resulting transformation is applied to both the Gaussian centers and scales, enabling accurate rendering and effective supervision, as illustrated in Figs.~\ref{fig:alignement}(c) and (d).



\begin{figure*}[t]
\centering
\includegraphics[width=1.0\textwidth]{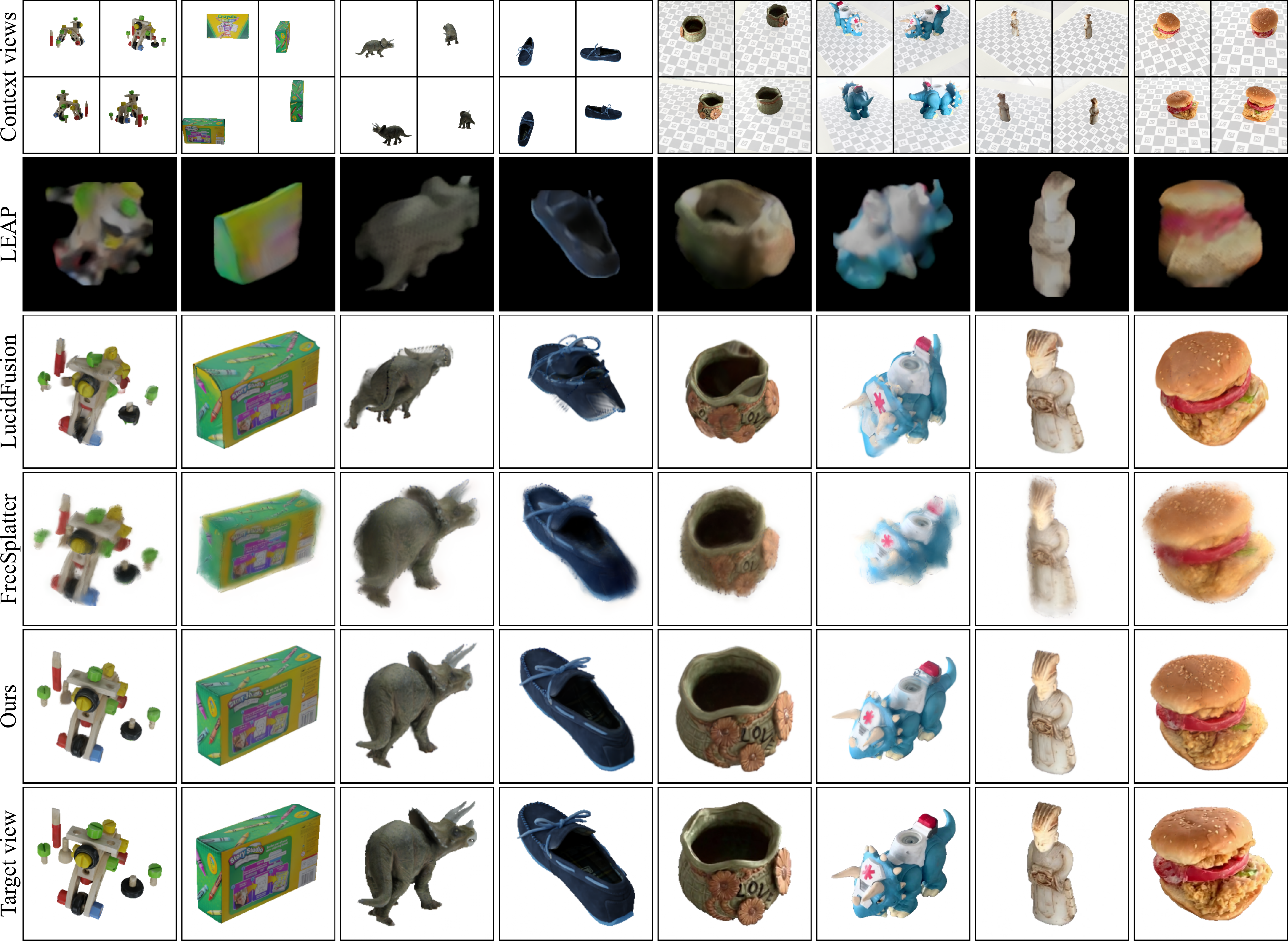}
\caption{Results of novel-view synthesis on the GSO and OmniObject3D dataset}
\label{fig:result_all}
\end{figure*}

\paragraph{Training loss.}

The training loss is computed between target images and those rendered from the aligned Gaussians.
It consists of a pixel-wise reconstruction loss (MSE) and a perceptual loss (LPIPS~\cite{Zhang2018}), weighted by 0.5 in all experiments.
\begin{table*}[tb]
\caption{Quantitative comparison in novel-view synthesis on the GSO and OmniObject3D dataset}
\label{tab:quantitative_comparison}
\centering
\begin{tabular}{lcccccc}
\toprule
 & \multicolumn{3}{c}{GSO} & \multicolumn{3}{c}{OmniObject3D} \\
& PSNR$\uparrow$ & SSIM$\uparrow$ & LPIPS$\downarrow$ & PSNR$\uparrow$ & SSIM$\uparrow$ & LPIPS$\downarrow$ \\
\midrule
LucidFusion & 14.966 & 0.820 & 0.229 & 14.206 & 0.798 & 0.249 \\
FreeSplatter w/o recentering & 12.475 & 0.815 & 0.265 & 12.145 & 0.796 & 0.288 \\
FreeSplatter w/ recentering & 16.698 & 0.825 & 0.232 & 15.807 & 0.803 & 0.245 \\
Ours & {\bf 22.910} & {\bf 0.883} & {\bf 0.110} & {\bf 18.861} & {\bf 0.816} & {\bf 0.182} \\
\bottomrule
\end{tabular}
\end{table*}

\section{Experiments}

\subsection{Dataset}

We adopt pretrained FreeSplatter~\cite{Xu2024} as our base model, originally trained on the Objaverse dataset~\cite{Deitke2023}, which contains 800K 3D assets. For training our method, we use the G-Objaverse dataset~\cite{Zuo2024}, a high-quality subset comprising 280K assets with multi-view images. Each scene in G-Objaverse is rendered from 38 favorable views. Following the protocol of~\cite{Chen2024lara}, we apply K-means clustering to divide camera positions into four clusters and sample one view from each cluster for 3D Gaussian labeling.

For evaluation, we use the Google Scanned Objects (GSO) dataset~\cite{Downs2022} and the OmniObject3D dataset~\cite{Wu2023}. From GSO, we select 50 scenes and render 32 unfavorable views per scene by applying random translations in arbitrary directions. Each view includes both RGB and depth maps. Of these, four views are used as context inputs, and the remaining 28 as target views. For OmniObject3D, we follow the preprocessed subset in~\cite{Yang2024}, selecting 20 scenes with provided foreground masks. The same four sparse views used in~\cite{Yang2024} are adopted as inputs, and 28 additional views are used as targets. These input views typically exhibit large camera baselines and off-center object placement, thus qualifying as unfavorable views as defined in Sec.~\ref{sec:define_ufv}.
Additionally, we apply 2D Gaussian splatting~\cite{Huang2024} to 200 frames in each scene to estimate depth maps, which are used for alignment of the estimated 3D Gaussians during evaluation.

\subsection{Implementation}
During training, we render four context and four target views ($N=4$) from the labeled Gaussians. All images are rendered at $512 \times 512$ resolution. We use spherical harmonics of degree $d = 1$, and the adapter operates with a patch size of $p = 8$. Training uses AdamW~\cite{Loshchilov2019} with $\beta_1 = 0.9$, $\beta_2 = 0.95$, a weight decay of 0.05, and a learning rate of $5.0 \times 10^{-5}$. Training is run for 40K iterations on four NVIDIA A100 GPUs with a per-GPU batch size of 2, completing in approximately 1.5 days.

\subsection{Experimental results}
\paragraph{Baselines.}
We compare our method primarily against FreeSplatter~\cite{Xu2024} and LucidFusion~\cite{He2025}, two state-of-the-art pose-free feed-forward 3DGS models trained on Objaverse for object-centric scenes, where training images are rendered from favorable viewpoints.
We also include LEAP~\cite{Jiang2024LEAP}, a pose-free feed-forward model with neural-volume-based representation trained on the Kubric-ShapeNet dataset~\cite{Jiang2024}, although we only perform qualitative comparison due to the difficulty of aligning generated images to the target views.  
PF-LRM~\cite{Wang2024} and SpaRP~\cite{Xu2024_sparp} are not included as their codes are not publicly available.

We evaluate FreeSplatter in two configurations: using the original sparse inputs (``w/o recentering'') and with recentered inputs (``w/ recentering''). For LucidFusion and LEAP, input images are resized to $256 \times 256$ and $224 \times 224$, respectively, to match their training settings.

\paragraph{Results.}

Figure~\ref{fig:result_all} presents qualitative comparisons on GSO and OmniObject3D datasets. 
Each sample includes the context views, ground-truth target, and outputs from LEAP, LucidFusion, FreeSplatter (w/ recentering), and our method. For all methods except LEAP, we apply our alignment strategy (Sec.~\ref{sec:training}); for LEAP, camera poses are manually tuned.

The results show that LEAP tends to produce overly smoothed images. 
LucidFusion struggles to recover accurate geometry under unfavorable views. 
FreeSplatter yields blurry outputs due to inconsistencies in Gaussian predictions across views. In contrast, our method effectively estimates Gaussians while maintaining multi-view consistency through adaptation, resulting in high-quality novel-view renderings. Please refer to the appendix for additional results.

Table~\ref{tab:quantitative_comparison} summarizes the quantitative results using PSNR, SSIM~\cite{Zhou2004}, and LPIPS~\cite{Zhang2018}. FreeSplatter w/o recentering performs poorly under unfavorable views, and recentering alone does not fully address this issue. 
LucidFusion also underperforms.
In contrast, our method significantly outperforms these baselines across all metrics, demonstrating effective adaptation to challenging viewpoints.

\subsection{Discussion}

\begin{table}[t]
  \centering
  \caption{Ablation study on each component for adaptation. ``PE'' denotes positional embedding.}
  \label{tab:ablation}
    \begin{tabular}{cccccc}
      \toprule
      LoRA & Adapter & PE & PSNR$\uparrow$ & SSIM$\uparrow$ & LPIPS$\downarrow$ \\
      \midrule
       & $\checkmark$ & $\checkmark$ & 19.749 & 0.853 & 0.165 \\
      $\checkmark$ &  &  & 20.898 & 0.865 & 0.131 \\
      $\checkmark$ & $\checkmark$ &  & 21.367 & 0.868 & 0.127 \\
      $\checkmark$ & $\checkmark$ & $\checkmark$ & {\bf 22.910} & {\bf 0.883} & {\bf 0.110} \\
      \bottomrule
    \end{tabular}
\end{table}

\begin{figure}[t]
  \centering
  \includegraphics[width=1.0\linewidth]{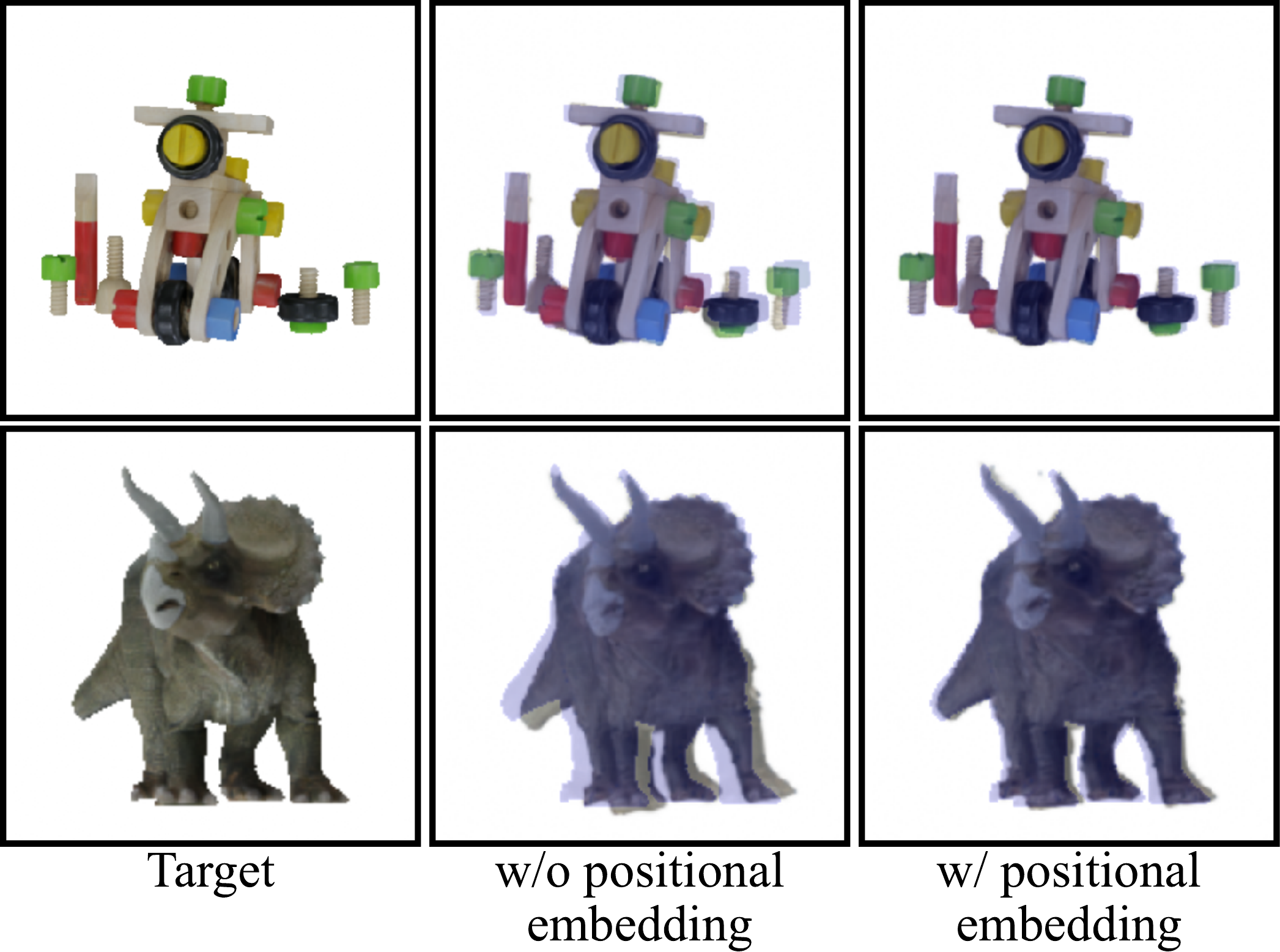}
  \caption{Target image and rendered novel-view images without and with positional embedding. Ground-truth targets are overlaid in blue. Positional embedding improves alignment and geometry fidelity.}
  \label{fig:ablation_pe}
\end{figure}

\paragraph{Ablation study on adaptation components.}
We evaluate each component in our adaptation framework (Tab.~\ref{tab:ablation}). The LoRA layers and the adapter each contribute to performance improvements. Adding the recentered positional embedding further improves consistency and geometry refinement, as illustrated in Fig.~\ref{fig:ablation_pe}.

\begin{figure}[t]
  \centering
  \includegraphics[width=1.\linewidth]{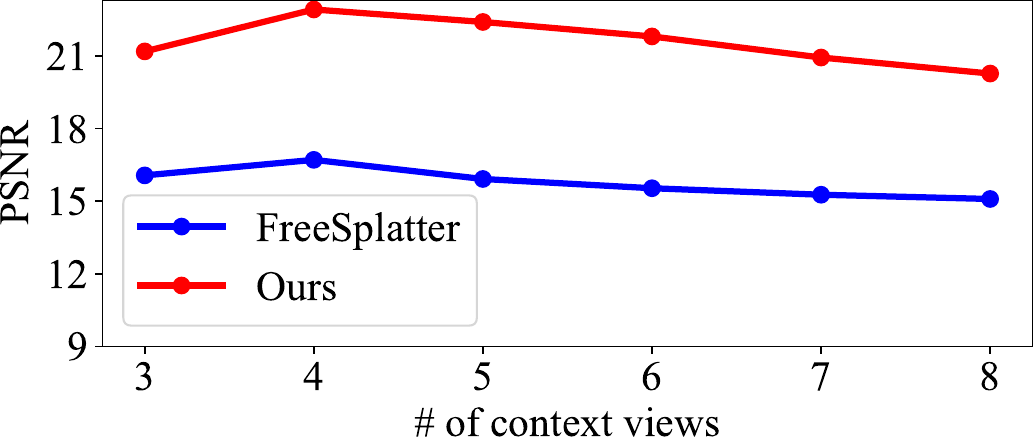}
  \caption{Performance comparison under different numbers of context views on the GSO dataset}
  \label{fig:plot_n_views}
\end{figure}

\paragraph{Number of input context views.}
Figure~\ref{fig:plot_n_views} shows performance under varying numbers of context views. As the number of unfavorable views increases, baseline models suffer due to noise accumulation in Gaussian estimation. In contrast, our method remains robust and maintains output quality despite large input variations.

\paragraph{Evaluation on favorable images.}
We also evaluate our method on favorable views in the GSO dataset, as shown in Tab.~\ref{tab:evaluation_favorable}. FreeSplatter performs well without recentering, but its performance drops significantly when the input images are recentered. Although our method slightly reduces PSNR on recentered favorable views, the LPIPS scores remain low, indicating that perceptual quality is well preserved (Fig. \ref{fig:challenges}(a)). These results suggest that our method retains visual fidelity even under favorable input conditions, supporting its general applicability when view conditions are unknown.


\paragraph{Limitation.}
As discussed in Sec.~\ref{sec:model}, our use of recentered input images prevents camera pose recovery via PnP-based methods. Developing a strategy to handle unfavorable views without relying on image recentering remains a promising direction for future work, which would further increase robustness in real-world applications requiring pose estimation.

Our method assumes object-centric scenes where the object is well-segmented. Extending the approach to handle more complex, scene-level environments with multiple objects and cluttered backgrounds remains an important direction for future work.

\begin{table}[t]
  \centering
  \caption{Evaluation on favorable images on the GSO dataset. ``RC'' denotes recentering.}
  \label{tab:evaluation_favorable}
    \begin{tabular}{lcccc}
      \toprule
       & RC & PNSR$\uparrow$ & SSIM$\uparrow$ & LPIPS$\downarrow$ \\
      \midrule 
      FreeSplatter & & {\bf 24.082} & {\bf 0.892} & 0.113 \\
      FreeSplatter & $\checkmark$ & 17.148 & 0.830 & 0.221 \\
      Ours & & 20.523 & 0.857 & 0.146 \\
      Ours & $\checkmark$ & 23.332 & 0.887 & {\bf 0.105} \\
      \bottomrule
    \end{tabular}
\end{table}

\section{Conclusion}
In this paper, we proposed a method to adapt a pretrained pose-free feed-forward 3DGS model to handle unfavorable input views.
Our approach introduces an adapter module that refines Gaussian predictions derived from recentered images, along with a Gaussian alignment strategy that enables accurate supervision by rendering target images.
We also presented a novel training framework that leverages only favorable-view images from an off-the-shelf dataset, thereby eliminating the need to create additional datasets.
Experimental results demonstrate that our method significantly improves performance under challenging conditions and enables high-quality novel-view synthesis.
This capability enhances the practical applicability of pose-free 3DGS models in real-world scenarios with unknown and varying camera poses.
{
    \small
    \bibliographystyle{ieeenat_fullname}
    \bibliography{main}
}
\clearpage
\setcounter{page}{1}
\maketitlesupplementary

\section{Details of architecture}

\subsection{LoRA layers}
Table \ref{tab:ablation_lora} presents a comparison of different configurations for inserting LoRA layers~\citep{Hu2022} on the GSO dataset.
All LoRA layers are configured with a rank of 64.
Since a large portion of the model consists of self-attention blocks, inserting LoRA layers into these blocks significantly affects accuracy. 
Among the configurations, applying LoRA to both the patchify convolution and the self-attention blocks yields the highest performance, and we adopt this setting throughout our experiments.

\begin{table}[h]
\caption{Ablation study of different configurations for inserting LoRA layers in the model. ``SA'' denotes self-attention blocks. Results are reported on the GSO dataset.}
\label{tab:ablation_lora}
\centering
\begin{adjustbox}{width=\linewidth}
\begin{tabular}{cccccc}
\toprule
Patchify & SA blocks & Linear head & PNSR$\uparrow$ & SSIM$\uparrow$ & LPIPS$\downarrow$ \\
\midrule
$\checkmark$ & & & 20.100 & 0.856 & 0.157 \\
& & $\checkmark$ & 19.917 & 0.854 & 0.163   \\
& $\checkmark$ & & 22.789 & 0.882 & 0.110   \\
$\checkmark$ & $\checkmark$ & &  {\bf 22.910} & {\bf 0.883} & {\bf 0.110}  \\
 & $\checkmark$ & $\checkmark$ & 22.877 & {\bf 0.883} & {\bf 0.110}   \\
$\checkmark$ & $\checkmark$ & $\checkmark$ &  22.819 & 0.882 & 0.111   \\
\bottomrule
\end{tabular}
\end{adjustbox}
\end{table}

\subsection{Adapter}
The adapter module receives the 3D Gaussian features $\{ \mathbf{G}_i\}_{i=1}^N$ and outputs residuals $\{ \Delta \mathbf{G}_i \}_{i=1}^N$. 
Each $\mathbf{G}_i \in \mathbb{R}^{H \times W \times (11+3(d+1)^2)}$ is first reshaped into patch tokens of size $H/p \times W/p \times p^2 (11+3(d+1)^2) $, where $p$ is the patch size. These are then linearly projected into tokens $\mathbf{t}_i \in \mathbb{R}^{H/p \times W/p \times D}$. 
The embedded tokens are added to the recentered 2D positional embedding $\overline{\mathbf{e}}_i$ as follows:
\begin{equation}
\mathbf{f}_i = \mathbf{t}_i + \overline{\mathbf{e}}_i.
\end{equation}
The resulting tokens $\{\mathbf{f}\}_{i=1}^N$ from all $N$ input images are concatenated and flattened into a single tensor $\mathbf{F}_0 \in \mathbb{R}^{NHW/p^2 \times D}$, which is processed by two transformer blocks:

\begin{align}
\mathbf{F}_j' &= \mathbf{F}_j + {\rm SelfAttention}({\rm LayerNorm}(\mathbf{F}_j)), \\
\mathbf{F}_{j+1} &= \mathbf{F}_j' + {\rm MLP}({\rm LayerNorm}(\mathbf{F}'_j)).
\end{align}
Here, ${\rm LayerNorm}$ is layer normalization, ${\rm SelfAttention}$ is multi-head self-attention, and ${\rm MLP}$ is a two-layer feedforward network with GELU activation.

The final output $\mathbf{F}_2$ is passed through a linear head to produce outputs of shape $NHW/p^2 \times p^2 (11+3(d+1)^2)$, which are reshaped to yield residuals $\{ \Delta \mathbf{G}_i\}_{i=1}^N$, with each $\Delta \mathbf{G}_i \in \mathbb{R}^{H \times W \times (11+3(d+1)^2)}$.
The linear head is initialized with zero weights, so the residuals are initially zero.

For the Gaussian centers, a $\tanh$ activation is applied and scaled by 0.05 to constrain translations within $[-0.05, 0.05]$ per axis. These residuals are added to the initial Gaussians before activation functions in the rasterizer~\citep{Kerbl2023}, ensuring compatibility with expected parameter ranges.

\section{Rendering details for unfavorable views}

To train the adapter, we label Gaussians for each sample from the G-Objaverse dataset~\citep{Zuo2024}, which contains only favorable-view images. Using these labeled Gaussians, we render both favorable target views and unfavorable context views.

For each rendering, we first sample a camera field of view $\theta_{\rm fov}$ from the range $[\ang{20}, \ang{50}]$. The image resolution is fixed at $(H, W) = (512, 512)$, and the focal length is computed as $f = H / (2 \tan(\theta_{\rm fov}/2))$. A base radius is defined as $r = 1 / \tan(\theta_{\rm fov}/2)$.
For each scene, we compute the centroid of the Gaussians and then sample four cameras on a sphere centered at this point with radius $r$. These cameras are oriented toward the centroid, with elevation angles randomly sampled from $[\ang{0}, \ang{45}]$. The azimuth angle of the first camera, $\theta_0$, is sampled from $[\ang{0}, \ang{90}]$, and the remaining three azimuth angles are offset $90^\circ$ apart.
Unfavorable context views are generated similarly, but with additional translations applied to each camera. The translation along each axis $(x, y, z)$ is uniformly sampled from the range $[-0.1r, 0.1r]$. The order of the rendered context views is shuffled during training.

\begin{figure*}[p]
  \centering
  \subfloat[]
  {\includegraphics[width=0.4\linewidth]{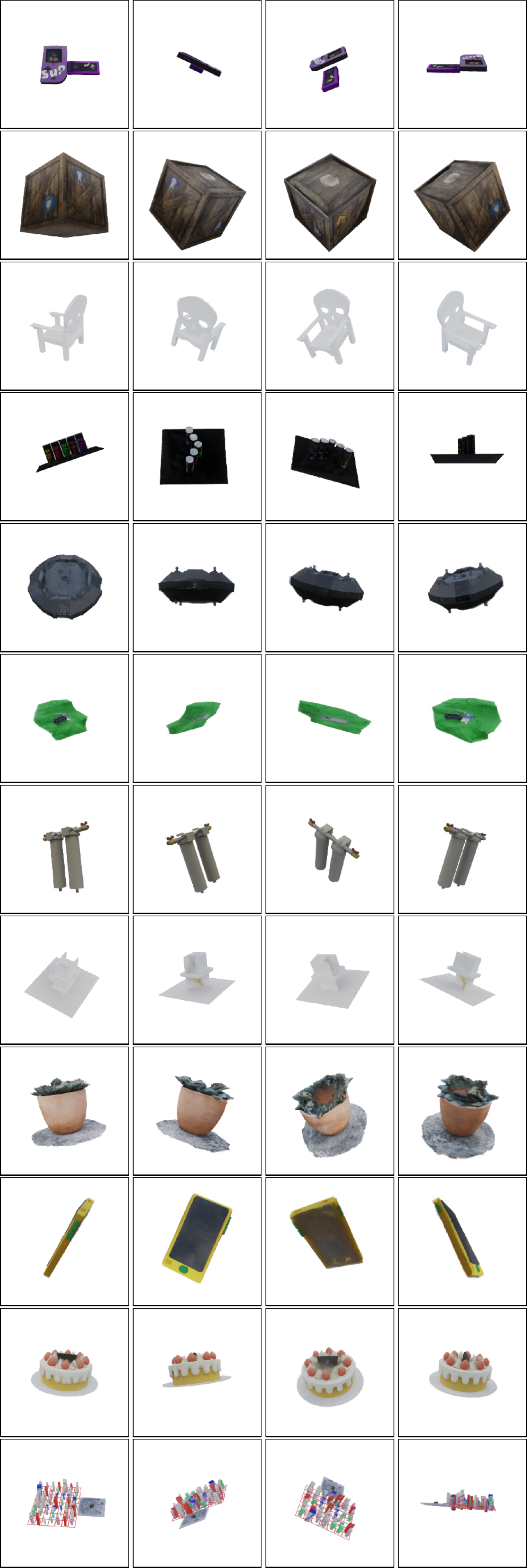}}
  \,
  \subfloat[]{\includegraphics[width=0.4\linewidth]{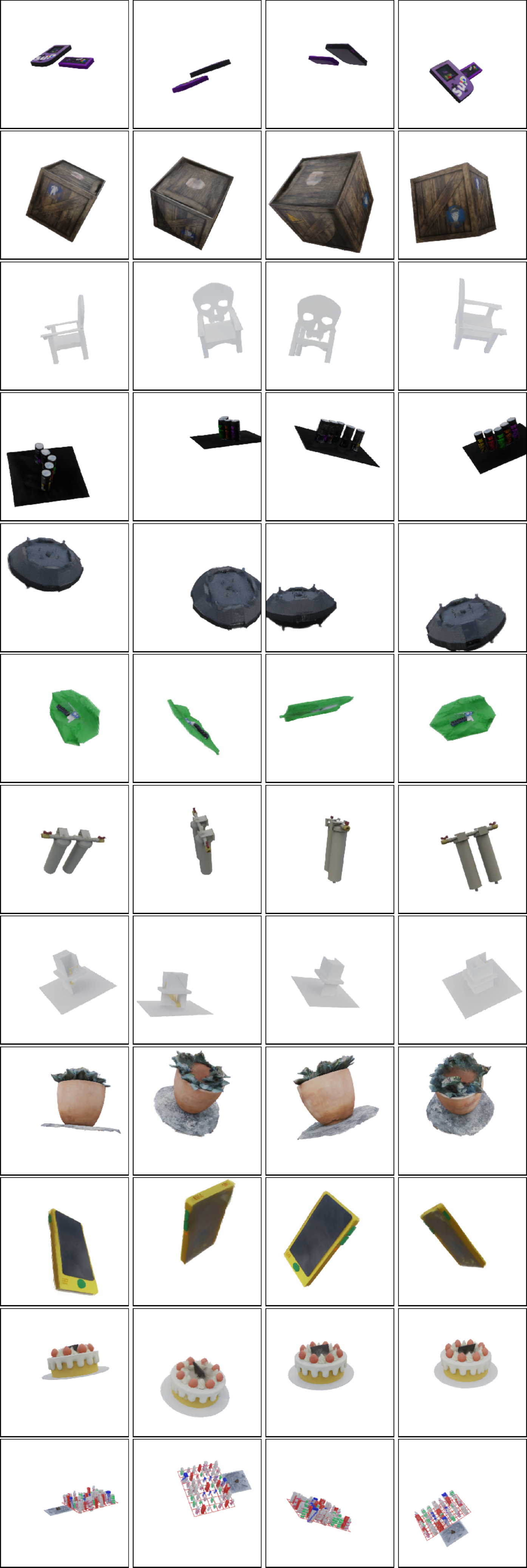}}
  \caption{Examples of rendered images used for training. (a) Favorable target images. (b) Unfavorable context images}
  \label{fig:example_images}
\end{figure*}

Figure~\ref{fig:example_images} illustrates examples of rendered images, where (a) shows favorable target views and (b) shows unfavorable context views.

\section{Closed-form solution for alignment}
As described in Sec.~\ref{sec:training}, we perform a linear alignment of the refined 3D Gaussians to enable rendering of target views during training. This is formulated as a weighted least squares problem:
\begin{gather}
\min_{a,\mathbf{b}} \sum_{i=1}^{N} \sum_{q \in \Omega} \overline{\mathbf{W}}_i^c(q)\left\| a
\widetilde{\boldsymbol{\mu}}_i(q)
+ \mathbf{b} - \overline{\mathbf{P}}_i^c(q)
\right\|^2,\\ \Omega = [1,W] \times [1,H]. \label{eq:appendix_alignment}
\end{gather}

Rewriting Eq.~(\ref{eq:appendix_alignment}), we obtain:
\begin{equation}
\min_{a,\mathbf{b}} \sum_{i=1}^{N} \sum_{q \in \Omega} \overline{\mathbf{W}}_i^c(q)\left\| \mathbf{X}_i(q)
\begin{bmatrix}
a \\
\mathbf{b}
\end{bmatrix}
- \overline{\mathbf{P}}_i^c(q)
\right\|^2,
\end{equation}
where $\mathbf{X}_i(q) = [\widetilde{\boldsymbol{\mu}}_i(q), \mathbf{J}] \in \mathbb{R}^{3 \times 4}$ and $\mathbf{J} \in \mathbb{R}^{3 \times 3}$ is the identity matrix. Differentiating with respect to $[a,\mathbf{b}^\top]^\top$ and solving yields:
\begin{align}
\begin{bmatrix}
a \\
\mathbf{b}
\end{bmatrix}
=& \left(\sum_{i=1}^{N} \sum_{q \in \Omega} \overline{\mathbf{W}}_i^c(q) \mathbf{X}_i(q)^\top \mathbf{X}_i(q) \right)^{-1}&\\
&\left( \sum_{i=1}^{N} \sum_{q \in \Omega} \overline{\mathbf{W}}_i^c(q) \mathbf{X}_i(q)^\top \bar{\mathbf{P}}_i^c(q)\right).&
\end{align}

Using the estimated $a$ and $\mathbf{b}$, we transform the Gaussian center and scale as follows:
\begin{align}
\widetilde{\boldsymbol{\mu}}_i(q) & \leftarrow a \widetilde{\boldsymbol{\mu}}_i(q) + \mathbf{b}, \\
\widetilde{\mathbf{s}}_i(q) & \leftarrow a \widetilde{\mathbf{s}}_i(q).
\end{align}

\section{Additional results}

\begin{figure*}[p]
  \centering
  \includegraphics[width=0.8\linewidth]{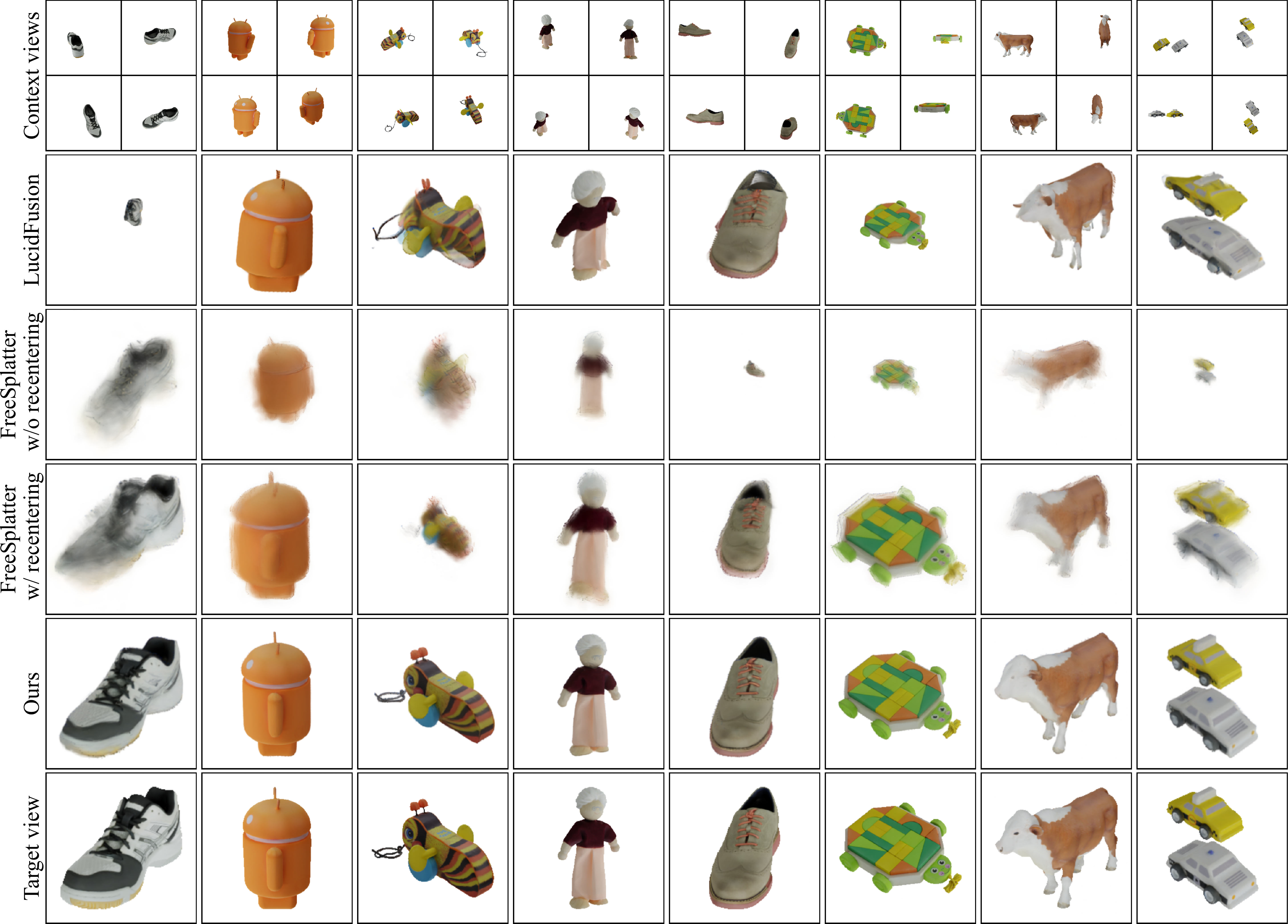}
  \caption{Additional results of novel-view synthesis on the GSO dataset}
  \label{fig:gso_additional}
\end{figure*}

\begin{figure*}[p]
  \centering
  \includegraphics[width=0.8\linewidth]{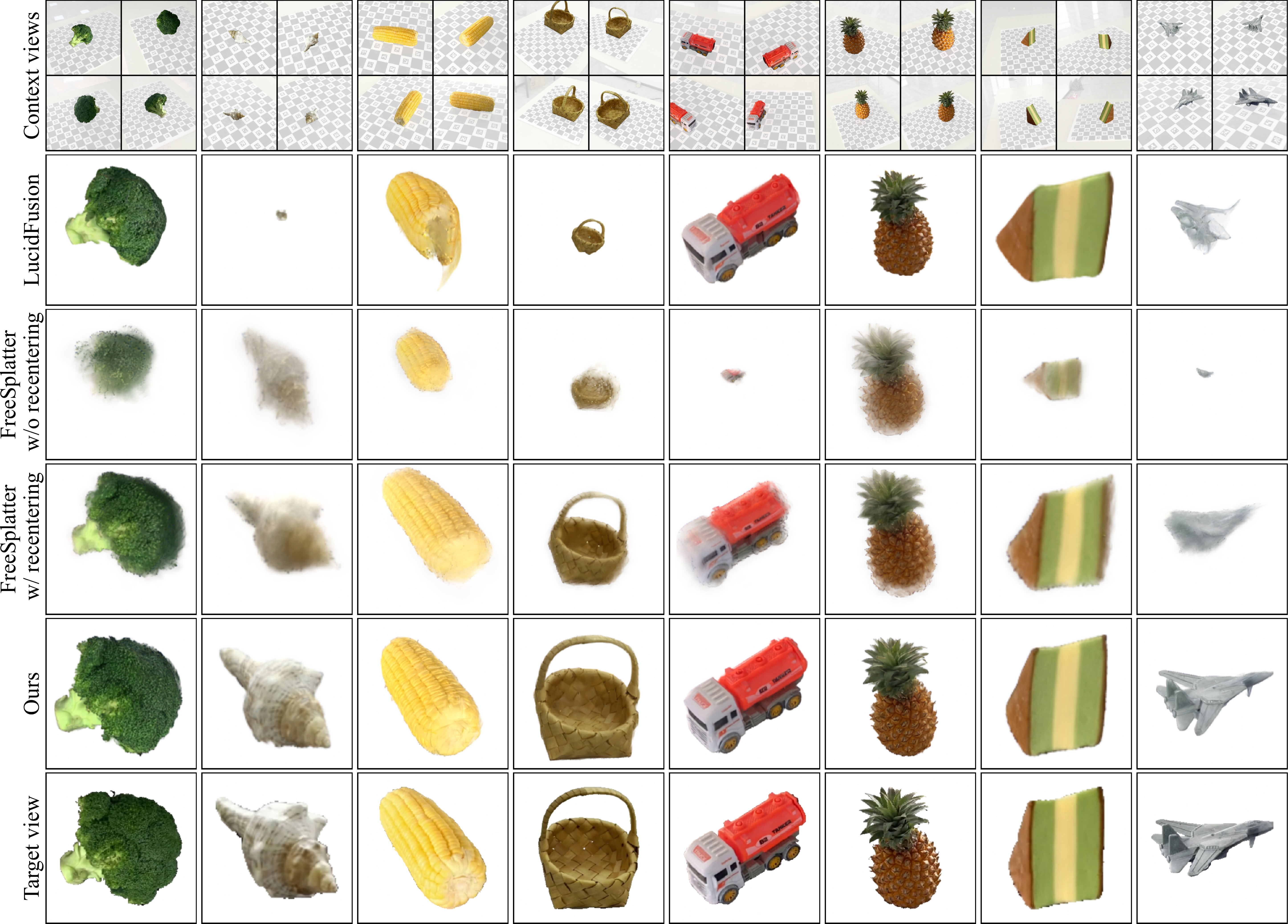}
  \caption{Additional results of novel-view synthesis on the OmniObject3D dataset}
  \label{fig:omni3d_additional}
\end{figure*}

Figures~\ref{fig:gso_additional} and~\ref{fig:omni3d_additional} show additional qualitative comparisons on the GSO and OmniObject3D datasets.
Each example includes the context views, a ground-truth target view, and novel-view results from FreeSplatter (w/ and w/o recentering)~\citep{Xu2024}, LucidFusion~\citep{He2025}, and our method. The alignment method described in Sec.~\ref{sec:training} is applied for rendering target views.
Our method consistently produces sharper and more reliable outputs under challenging unfavorable conditions.

\clearpage

\clearpage

\end{document}